# Toward Neuromic Computing:

# Neurons as Autoencoders

Larry Bull

Computer Science Research Centre

University of the West of England, Bristol BS16 1QY UK

Larry.Bull@uwe.ac.uk

**Abstract**

This short paper presents the idea that neural backpropagation is using dendritic processing to enable individual neurons to perform autoencoding. Using a very simple connection weight search heuristic and artificial neural network model, the effects of interleaving autoencoding for each neuron in a hidden layer of a feedforward network are explored. This is contrasted with the equivalent standard layered approach to autoencoding. It is shown that such individualised processing is not detrimental and can improve network learning.

**Introduction**

The branches of dendritic trees act as separate subunits with individual activation processing capabilities before their overall activity is integrated by the cell's soma and passed on (e.g., (Koch et al., 1982)). This has led to a number of computational models of their behaviour, typically with the synapse signal processing cast as feedforward, multi-layered networks. Initial work used networks consisting of two layers, the first as the dendrites and the second as the soma (Poiraz et al., 2003) but larger, deeper models have since been introduced (see (Poirazi & Papoutsi, 2020) for a review). Many neurons generate a backwards "echo" of their activation which passes up the dendritic tree. Such neural backpropagation has a number of competing explanations around synaptic plasticity and long-term potentiation (e.g., see (Water et al., 2004)). In their basic form, autoencoders learn efficient encodings of data by encoding the provided input to a (typically) lower dimension and then decoding it, ideally with minimal loss (e.g., see (Hinton & Salakhutdinov, 2006) for an early example). It is here suggested that neural backpropagation enables individual cells to perform autoencoding and therefore identify and pass on underlying structure(s) in the information flowing into and created by the brain. Figure 1 shows how autoencoders can be mapped to a single neuron.

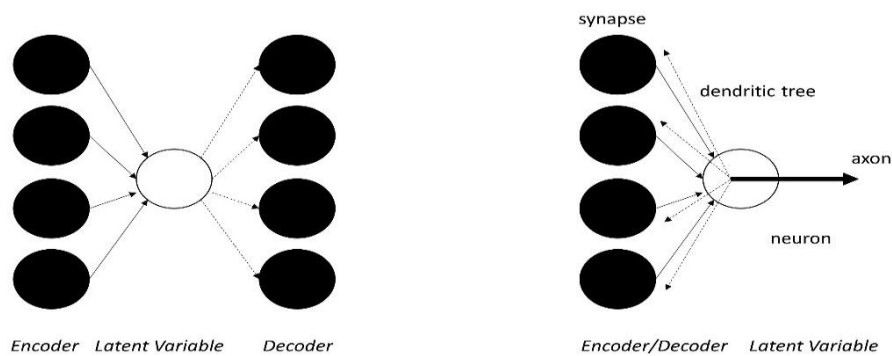

Figure 1: Showing how neural backpropagation (right) can be seen as a form of autoencoding (left).

The use of simple autoencoding capabilities by individual neurons within a network is explored here.

**Tuneable Data from the NK Model**

The well-known NK model (Kauffman & Weinberger, 1989) is used here to provide a flexible source of data for a regression task. In the NK model the features of the genome/system are specified by two parameters: $N$, the length of the genome; and $K$, the number of genes that have an effect on the fitness contribution of each (binary) gene. Thus increasing $K$ with respect to $N$ increases the epistatic linkage, increasing the ruggedness of the fitness/problem landscape. The increase in epistasis increases the number of optima, increases the steepness of their sides, and decreases their correlation. The model assumes all intragenome interactions are so complex that it is only appropriate to assign random values to their effects on fitness. For each of the possible $K$ interactions a table of $2^{(K+1)}$ rows is created with all entries in the range 0.0 to 1.0, such that there is one fitness for each combination of traits (Figure 2). The fitness contribution of each gene is found from the table. These fitnesses are then summed and normalized by $N$ to give the selective fitness of the total genome. The NK model can be used as a flexible regression function for machine learning (Bull, 2022): by altering $K$, the degree of interdependence between the features (genes) in the binary input space is tuneable, as is the number of features by altering $N$. Hence the conditions under which an algorithm may be beneficial – or not – can be teased out by systematically altering $N$ and $K$, alongside standard training scheme variations, etc.

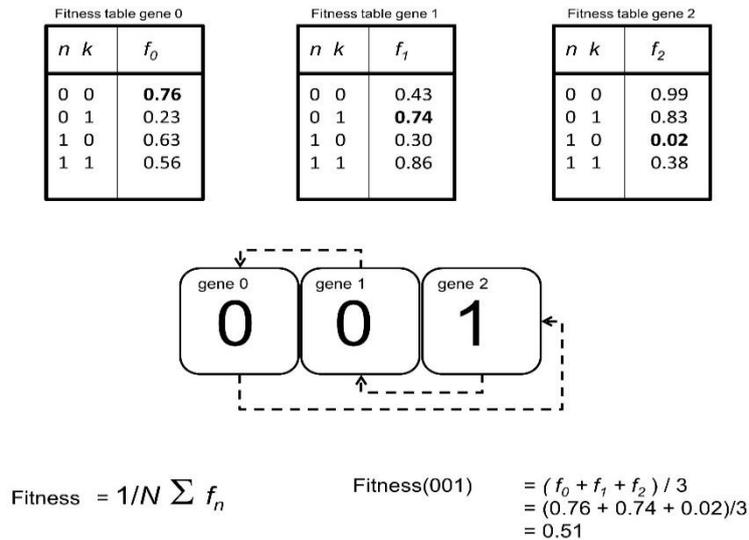

Figure 2: An example NK model where *N*=3 and *K*=1.

**Neuron Autoencoding**

A single dendritic branch can connect with many synapses and potentially exhibit a variety of computational processes (e.g., see (London & Häusser, 2005)). A greatly simplified scheme is adopted here of one synapse connection per branch (Figure 1). Standard multi-layered perceptrons (MLP) (Rosenblatt, 1961) are used, with each represented by the weights for a fully-connected, two-layered network. Each MLP has $N$ input nodes, $H$ hidden layer nodes, and one output node, with all nodes containing a bias and using a sigmoid transfer function (as opposed to the original linear activation). At the synaptic level, the realisation of accurately sharing error signals between neurons as in neural network backpropagation remains unclear (e.g., see (Lillicrap et al., 2020)). For simplicity, the weights are here learned using a random hill-climber starting with an initially randomly generated MLP. Weights are seeded in the range [-1.0,1.0] and a learning cycle consists of one randomly chosen weight from either the hidden layer or output layer being adjusted by a random amount from the range [-$R$,+$R$]. Adjusting a hidden layer node weight is here termed an autoencoding cycle and adjusting an output layer node is here termed a task learning cycle, since the former is minimising decoder error as its objective and the latter is minimising regression function error as its objective. Whether a learning cycle is an autoencoding or task cycle is determined at random with equal probability (0.5).

To include neuron-level autoencoding each hidden layer neuron also has *N* extra weights for their decoder layer, seeded in the same way as other weights (Figure 3). An autoencoding cycle occurs when a hidden layer node is chosen for adjustment. Performance is measured as the mean squared error between all *N* input values and their corresponding decoded output for the neuron experiencing the change over the training set. A task learning cycle occurs when an output layer node is chosen for alteration, with performance measured as the mean squared error over the training set and the MLP's output as in a standard supervised scenario. In both cases, the new network configuration is chosen if its mean squared error is reduced in comparison to that of the current solution, with ties broken at random. For comparison, an equivalent layer-based autoencoder is created: an extra *N* nodes are associated with the hidden layer and the accuracy is determined as described for a single neuron autoencoder (Figure 4, left).

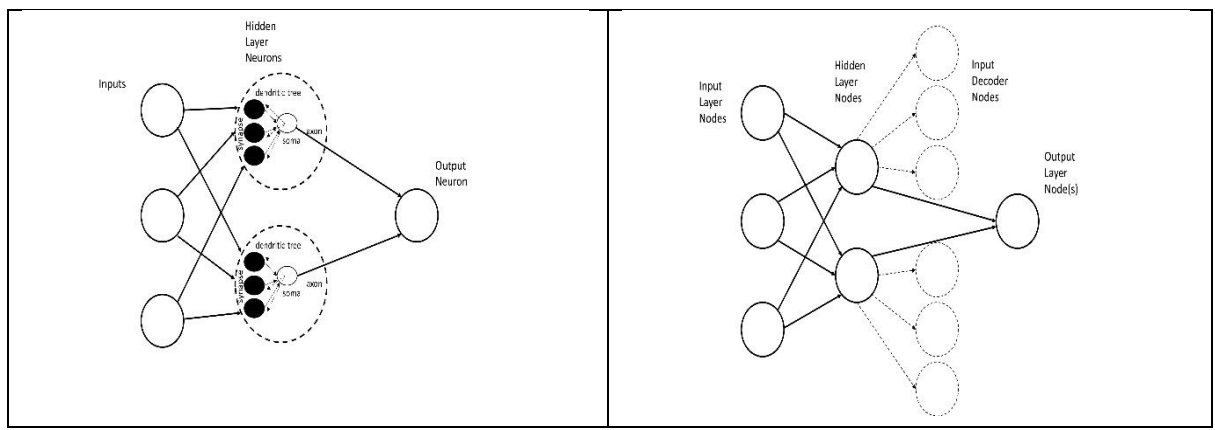

Figure 3: Showing the general schematic for networks containing autoencoding neurons in the hidden layer (left) and the resulting structure for the networks used to implement that here in which each hidden layer node is connected to both the next layer and its own autoencoder (right), termed neuron autoencoder networks (NAN).

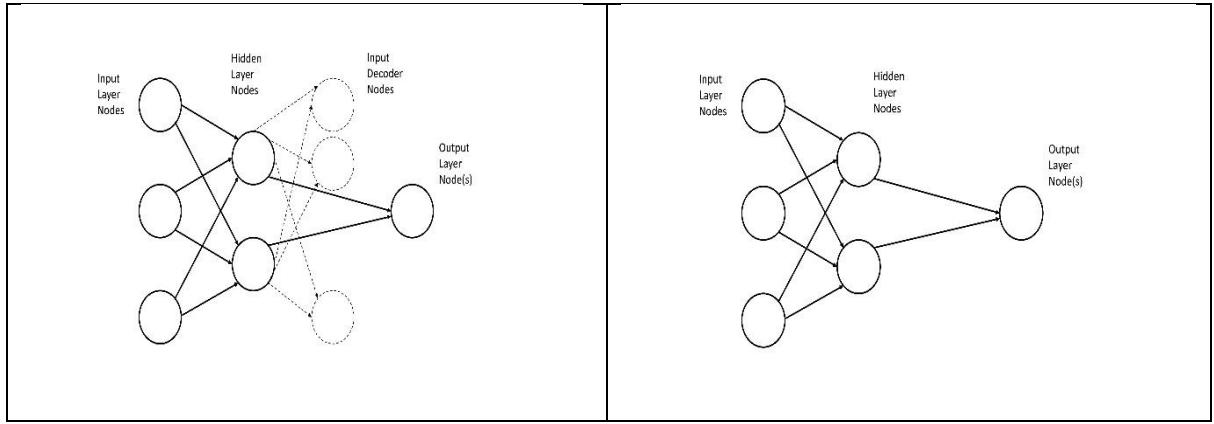

Figure 4: Showing the more standard approach of one decoder layer (left), autoencoder neural networks (ANN), and a basic feedforward network (right)(NN).

All results reported here are the average of twenty runs, each over ten thousand training iterations, with $H$=10, and $R$=1.0. Training and test sets of 1000 randomly created NK examples are used, with $0<K\leq15$, $20\leq N\leq1000$. For each binary genome/data point and its fitness/output, the typical use of -1.0 to replace zeroes as inputs is adopted. The Shapiro-Wilk test is used to confirm the normal distribution of results.

Figure 5 shows example training set errors over time for a given $N$ and $K$, both in terms of the overall regression task (left) and the autoencoding (right). As Figure 6 shows, for small $N$, neuron autoencoding proves beneficial for all $K$ tried (T-test, $p<0.05$) and there is no significant difference between the two for a larger $N$ (T-test, $p\geq0.05$). That is, the localised information compression potentially ocurring within dendritic trees is found to be beneficial, or at worst neutral, here.

In comparison to the equivalent purely supervised learning approach, both forms of autoencoding introduce extra connection weights and a less task-direct learning cycle half of the time. Figure 7 shows example results from using the same approach to train a standard MLP (Figure 4, right) which demonstrate a somewhat anticipated increase in learning speed for small $N$. However, MLP performance after 10,000 iterations is not significantly different to

NAN for small $N$ when $K>10$ and, for larger $N$, NAN are significantly better for all $K$ (T-test, $p<0.05$), where the learning speed benefit is also lost. That is, the localised autoencoding proves neutral for small, complex functions and beneficial for larger functions in general.

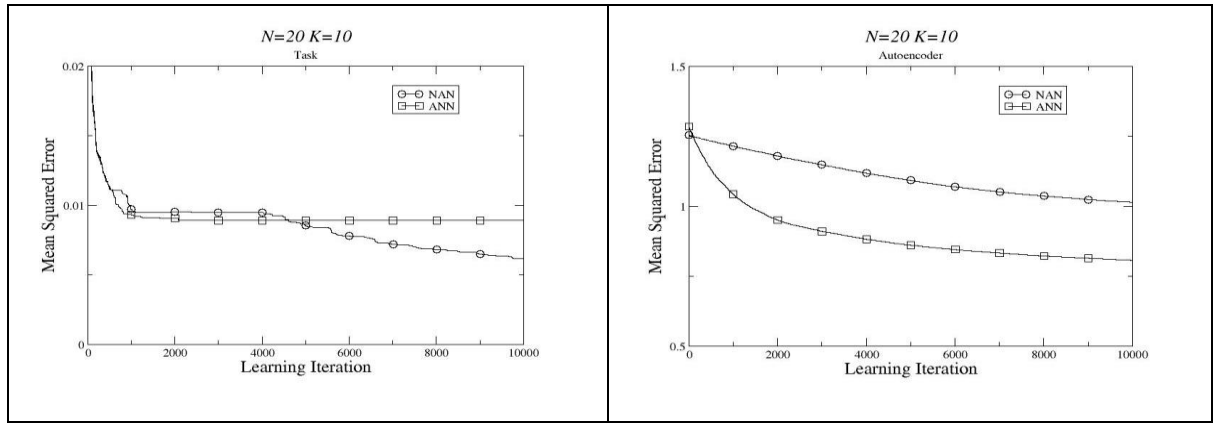

Figure 5: Showing the training errors of the neuron autoencoder networks (NAN) and the autoencoding neural networks (ANN), on an example dataset. The NAN autoencoder error is the average of the $H$ hidden layer nodes'.

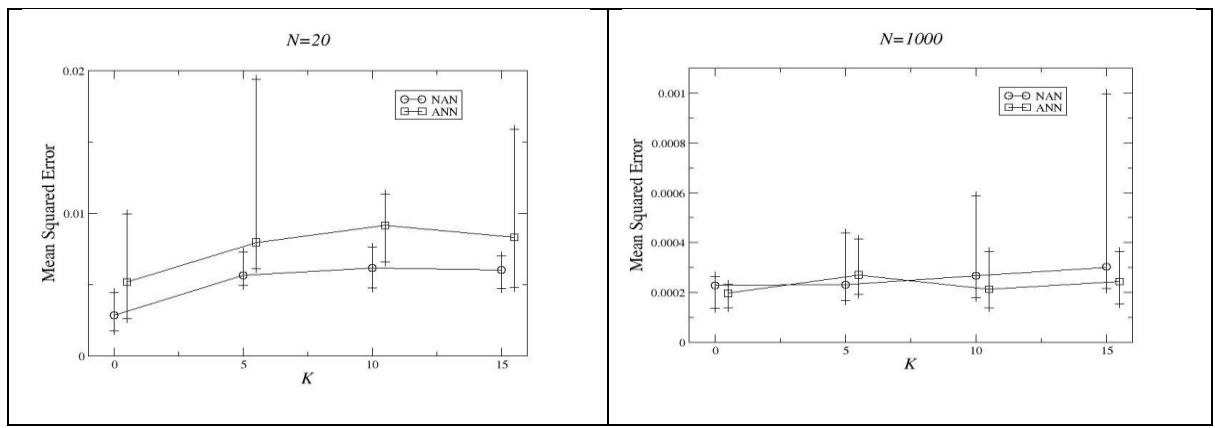

Figure 6: Showing the test errors of the best solution reached after 10,000 iterations, on datasets with varying feature interdependence ($K$) and number of features ($N$), by networks with neuron (NAN) and hidden layer (ANN) autoencoding. Error bars show min and max values.

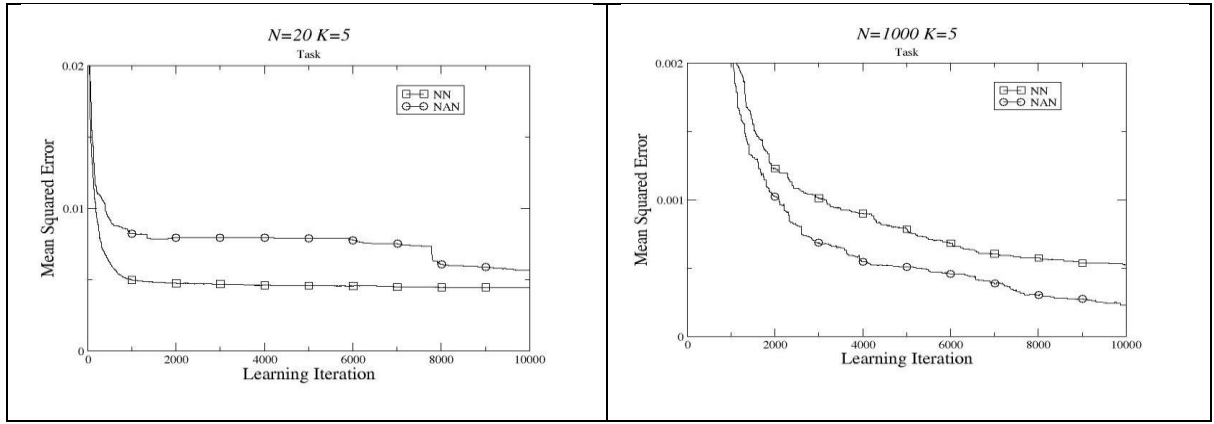

Figure 7: Showing example test errors during learning of neuron autoencoder networks (NAN) in comparison to equivalent standard neural networks (NN) for different *N* and *K*=5.

**Conclusion**

It has been suggested here that individual neurons can act as autoencoders. This both provides a new explanation for neural backpropagation and identifies a neuron-specific local optimisation function which is likely to be globally beneficial in a number of ways. It recasts the need for sharing accurate error signals between cells as rewarding the assembly of appropriate latent variable processing which may be facilitated through simple, well-established (Hebbian-like) mechanisms. That is, the basic computational unit being brought in to a circuit constructing generalizations is itself a compact description of an aspect(s) of the problem space. Individual neurons may be able to exhibit generative behaviour through synapses on the soma, assuming the usual control of the possible positive feedback is relaxed. As things like brain region reassignment after limb loss and *in vitro* cell culturing demonstrate, neurons appear to constantly seek each other out: they balance optimising the accuracy of their latent variable with maximising the number of synaptic connections they have (or are actively processing) at any time. Numerous neuron-specific mechanisms have been identified within the various omics disciplines which underpin the context-sensitive and wide scale plasticity over multiple timescales seen in natural systems – mRNA, transposons, methylation, etc. All such self-modification can be locally evaluated/controlled under autoencoding accuracy and future work will seek to include some of these processes – what is here termed neuromic computing.

A very simple learning process has been used. Dendrites exploit a myriad of ion channels, structural features, local mRNA, local protein synthesis, etc. in their processing. Future work will consider in more biological detail how an error-based learning gradient between the forward passing synaptic activations and the returning axon activation might be approximated to facilitate autoencoding in a neuron, such as through NMDA conductance (e.g., after (Bicknell & Häusser, 2021)). Moreover, the basic architecture has the potential to exploit (asynchronous) parallel learning both within and between layers.

Use of autoencoder neurons within other network architectures and on other benchmark tasks is currently being explored.